\documentclass[10pt,twocolumn,letterpaper]{article}
\usepackage[accsupp]{axessibility}
\usepackage{iccv}
\usepackage{times}
\usepackage{epsfig}
\usepackage{graphicx}
\usepackage{amsmath}
\usepackage{amssymb}
\usepackage{caption}
\usepackage[table,xcdraw]{xcolor}
\usepackage{booktabs}
\usepackage{textcmds}
\usepackage[autostyle]{csquotes}
\usepackage{authblk}

\usepackage[breaklinks=true,bookmarks=false]{hyperref}

\iccvfinalcopy 

\def\iccvPaperID{****} 

\ificcvfinal\fi

\begin{document}

\title{Understanding of Emotion Perception from Art}

%
\author[1]{Digbalay Bose}
\author[1]{Krishna Somandepalli}
\author[1]{Souvik Kundu}
\author[1]{Rimita Lahiri}
\author[1,2]{Jonathan Gratch}
\author[1]{Shrikanth Narayanan}
\affil[1]{University of Southern California, Los Angeles, CA}
\affil[2]{USC Institute of Creative Technologies}
\affil[1]{\tt {{\small\{dbose@,somandep@,souvikku@,rlahiri@,shri@ee.\}usc.edu}}}
\affil[1,2]{\tt {{\small\{gratch@ict.\}usc.edu}}}
\maketitle

\ificcvfinal\thispagestyle{empty}\fi

\begin{abstract}
 Computational modeling of the emotions evoked by art in humans is a challenging problem because of the subjective and nuanced nature of art and affective signals. 
 In this paper, we consider the above-mentioned problem of understanding emotions evoked in viewers by artwork using both text and visual modalities.  Specifically, we analyze images and the accompanying text captions from the viewers expressing emotions as a multimodal classification task. Our results show that single-stream multimodal transformer-based models like MMBT and VisualBERT perform better compared to both image-only models and dual-stream multimodal models having separate pathways for text and image modalities. We also observe improvements in performance for extreme positive and negative emotion classes, when a single-stream model like MMBT is compared with a text-only transformer model like BERT.
\end{abstract}

\section{Introduction}

Visual art is a rich medium for expressing human thoughts and emotions. This can manifest in multiple diverse forms; for example, consider the  well-known works such as the dreamy starry night canvas created by Van Gogh, the mythical figurines of Boticelli's Primavera or the mystical characters like Da Vinci's Mona Lisa. 
A common unifying factor of an artwork is its ability to convey and evoke emotional reaction in a viewer. In the affective computing realm, prior works like \cite{Van_Gogh_Affect}, \cite{LREC18-ArtEmo}, have explored computational models for understanding emotion elicited by art works. However, modeling emotional responses evoked by art works is especially challenged by the uncertainty associated with the subjectivity across individual viewers' felt experiences. Based on familiarity with art styles, formative experiences and their present mental state and mood, different persons might interpret the same artwork from multiple diverse perspectives. Since language is a powerful tool for explaining subjective emotional responses \cite{ortony1990cognitive},  one approach is to have  natural language explanations of \textit{why} people felt a certain emotion upon looking at a piece of art.






\begin{figure}[!t]
    \centering
    \includegraphics[width=\columnwidth]{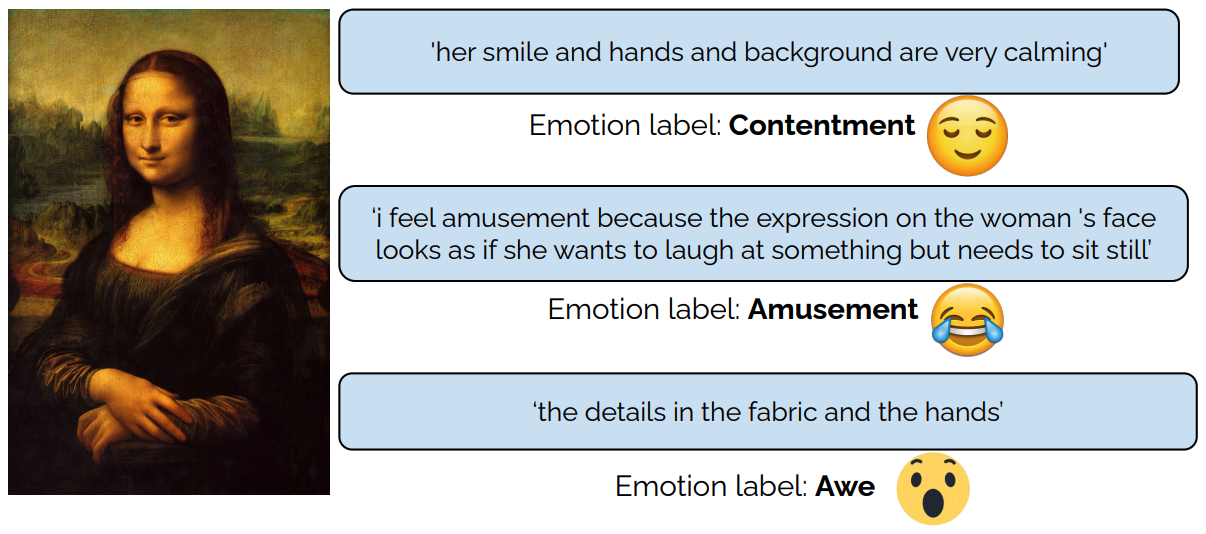}
    \caption{Example of different captions and emotion labels associated with Monalisa painting. Image taken from \url{https://www.wikiart.org/} and the captions with their emotion labels are taken from Artemis dataset \cite{Artemis}.}
    \label{Monalisa sample}
\vspace{-5mm}
\end{figure}
\par 

In this work, we investigate the problem of modeling emotion perception from artwork both through the lens of text-only classification from the associated explanations and by considering multimodal interactions of image regions and the associated textual explanations. 
The experiments of this work 
are based on the recently released  Artemis dataset \cite{Artemis} that has 80K artwork images from WikiArt\footnote{https://www.wikiart.org/} along with 9 emotion class annotations. Artemis was considered because it includes around 400K subjective explanations regarding emotion labels associated with artwork. A sample artwork with associated emotional labels and explanations is shown in Fig 1. 


\section{Related work}
\label{sec:related}

\textbf{Image and video based emotion classification.} 
In the domain of modeling (categorical) affect evoked from art images, a system trained on the widely used stimuli set from the international affective picture system (IAPS) was evaluated on art masterpieces in \cite{yanulevskaya2008emotional}. 4000 artwork from the Wikiart repository were leveraged in \cite{LREC18-ArtEmo} and annotated for a range of emotions along with title and other aspects.

\par 
\textbf{Text based emotion classification.} 
Linguistic descriptions can be used in affective analysis such as reasoning about an expressed felt experience. A variety of computational text analysis methods and tools are available in the literature, building on a variety of word dictionary based resources such as LIWC\cite{Tausczik2010}
. 
With the advent of transformer based methods \cite{vaswani2017attention} in various natural language processing tasks, BERT \cite{devlin2019bert} based models have been explored for classifying emotion labels e.g., from Reddit comments  \cite{demszky2020goemotions}.

\par 
\textbf{Multimodal vision-language modeling.}
Recently transformer based models that use vision-language pretraining on large-scale vision datasets have been fine tuned for variety of downstream tasks. 
Examples of vision-language based models employing transformers for individual streams followed by fusion include LXMERT \cite{tan2019lxmert}, VILBERT \cite{lu2019vilbert}. Examples of single stream models based on concatenation of textual and visual inputs include MMBT \cite{kiela2019supervised}, VisualBERT \cite{li2019visualbert}. 

\section{Multimodal adaptation for evoked emotion understanding}
\label{sec:approach}



\begin{figure}[t!]
    \centering
    \includegraphics[width=0.95\columnwidth]{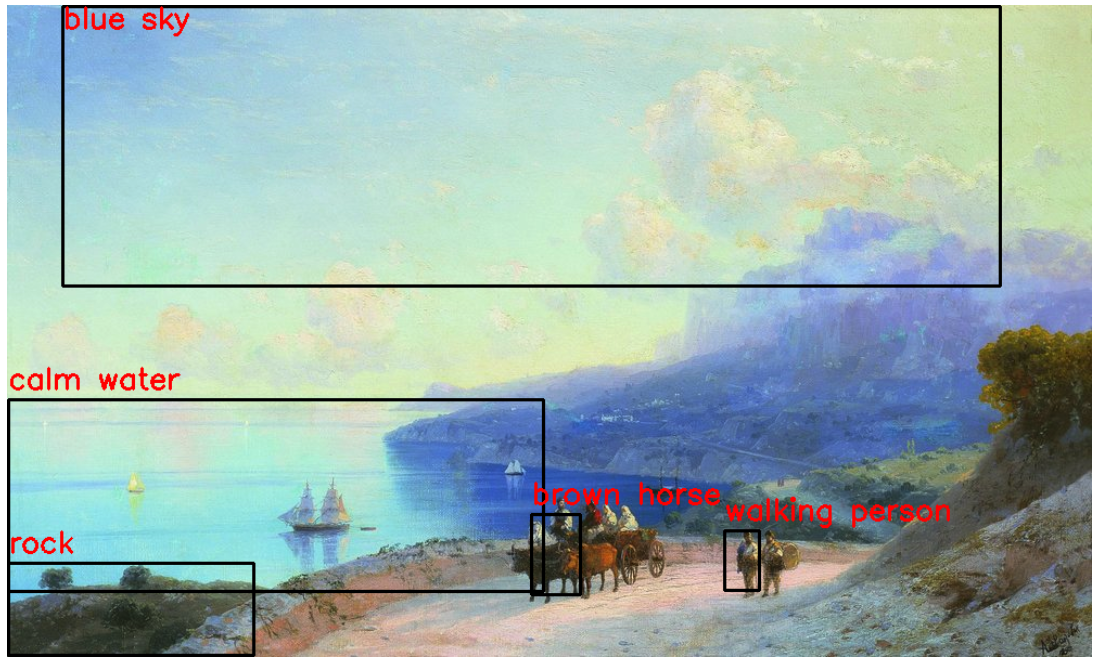}
    \caption{The caption associated with Ivan Aivazovsky's art work from Artemis Dataset is \textit{\qq{it makes me think of how things used to be and how simple life once was}}.  Image regions used for feature extraction are shown as black boxes with associated labels.}
    \label{calm surroundings}
\vspace{-5mm}
\end{figure}

Textual and visual information pertaining to evoked emotions associated with artwork contain complementary information: perceptually relevant affective cues in images, direct information in text descriptions about felt emotion.   
A trained text classifier model like BERT\cite{devlin2019bert} predicts sadness if only the caption \qq{\textit{it makes me think of how things used to be and how simple life once was}} is used. When the visual artwork image (Fig \ref{calm surroundings}) is shown, then the associated caption along with the artwork evokes a feeling of contentment. Hence we explore a series of multimodal model adaptations for evoked emotion prediction.

\subsection{Dual-stream models}
\begin{figure*}[h!]
    \centering
    \includegraphics[width=\textwidth]{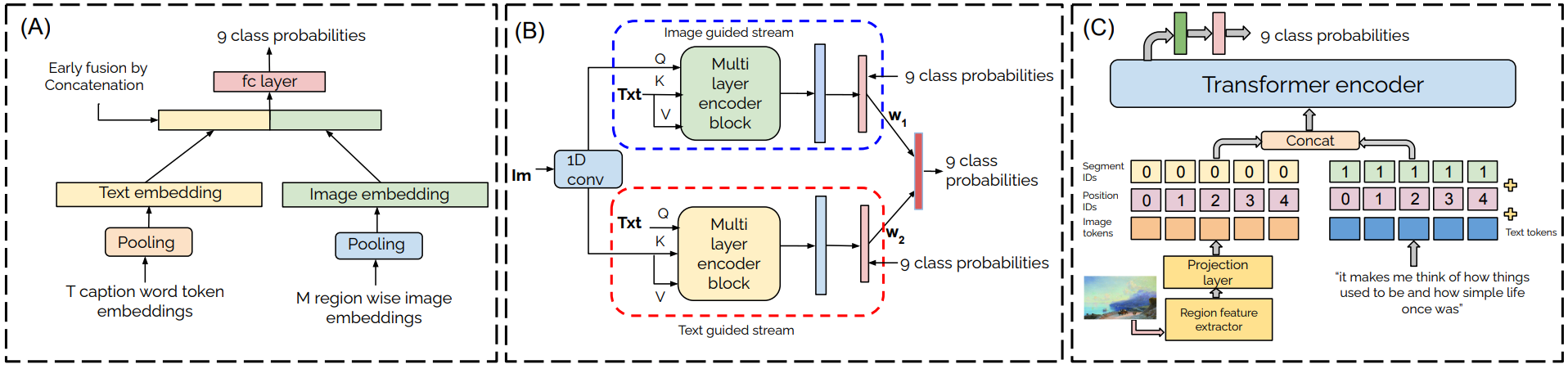}
    \caption{(A) Dual-stream early fusion model (\textbf{Early fusion avg pool}: Average of word token embeddings used as text embedding, \textbf{Early fusion first token}: First word token embedding used as pooled text embedding). (B) Dual-stream multi-layer encoder late fusion model (\textbf{Weighted late fusion}). Here Q: Query, K: Key, V: Value. Txt: Text caption embeddings, Im: Region wise image embeddings (C) Single-stream configurable MMBT model (\textbf{MMBT})}
    \label{Architectures}
\vspace{-4mm}
\end{figure*}

For the dual-stream model baseline, we pool the embeddings of text tokens associated with the captions and the region wise embeddings (avg-pooled) from image as shown in Fig.~\ref{Architectures}(A). The pooled embeddings are then concatenated followed by FC layers for emotion classification.
We further explore modeling of explicit interactions between captions and image region features by using image-guided and text-guided streams followed by a weighted late fusion scheme of predicted class probabilities from individual streams. As shown in Fig.~\ref{Architectures}(B),we use multi-layer self-attention based encoder block. The output embeddings from the encoder blocks of the respective streams are further average pooled and passed through fully connected (FC) layers for classification followed by fusion using weights $\mathbf{w_{1}}$ and $\mathbf{w_{2}}$. 


\subsection{Single-stream models}


For single-stream design of multimodal models, we rely on the multimodal bitransformer model (MMBT) \cite{kiela2019supervised} consisting of 12 encoder layers and include a configurable image encoder to extract region wise features, as shown in Fig.~\ref{Architectures}(C). Since the image and text features are combined together as a single input to the encoder model, separate segment IDs are provided to the image (segment ID = 0) and text (segment ID = 1) modalities. For individual modalities i.e., image and text, the input embeddings are obtained as the sum of token, positional and segment embeddings. The representation associated with [CLS] (classification) token is used for downstream emotion classification.


\vspace{-2.0 mm}

\section{Experiments and Results}
We conduct our experiments on the Artemis dataset \cite{Artemis}. Artemis is built on top of 81,446 art work images from the Wikiart dataset, spanning 27 art styles and 1119 artists from the 15th to 21st century. We perform the same train, test, and validation set split as in \cite{Artemis}. 
\vspace{-1 mm}
\subsection{Experimental Setup}
We use categorical cross entropy as loss function with label smoothing \cite{label_smoothing} as a regularization technique while training the multimodal models. We use the AdamW optimizer \cite{LoshchilovH19} and linear scheduler with warm-up (warm-up steps varied between 1000 to 10000). The learning rate is tuned in the range \{1e-5, 6e-5\} in steps of 1e-5. For text-based baselines, we fine-tune BERT \cite{devlin2019bert} to predict emotion classes from the text captions. For the multi-layer encoder design in dual stream weighted later fusion model, we use 5 encoder layers and 8 heads.
For the image modality, we extract bottom-up features \cite{Anderson2017up-down} from the top-$k$ salient image regions (sorted by scores) by using a Visual-Genome based pretrained Faster RCNN \cite{fastercnn} model with a ResNet-101\cite{He2015} backbone. We also look at  ResNeXt-152 C4 based object detection model \cite{zhang2021vinvl} to extract enhanced region-wise representations (VinVL).
For the image based models, we fine-tune VGG-16\cite{vgg16} and ResNet-50 \cite{He2015} pretrained on ImageNet \cite{ILSVRC15} by using a KL-divergence loss between the network outputs and per-image distribution of emotions based on responses of annotators.

\subsection{Results}

We compare performances in terms of 9 class accuracies and macro-F1 for image-only, text-only and multimodal models. After tuning on the validation set, the optimal weights for late fusion in the dual stream multi layer encoder model are obtained as $\mathbf{w_{1}}=0.76$ and $\mathbf{w_{2}}=0.24$ for the image- and text-guided streams respectively.
\begin{table}[h!]
\begin{tabular}{cccc}
\hline
\textbf{Model}                      & \textbf{Acc}            & \textbf{F1}                   & \textbf{Feat}                 \\ \hline
\multicolumn{4}{c}{\cellcolor[HTML]{DAE8FC}\textbf{Image (N = 79327)}}                                \\ \hline
VGG-16                     & 47.36         &      27.04                \\ \hline
ResNet-50                  & 44.98    & 21.31                            \\ \hline
\multicolumn{4}{c}{\cellcolor[HTML]{00D2CB}\textbf{Text (N = 429431)}}                                 \\ \hline
BERT                       & 66.2           & 61.42                &                      \\ \hline
\multicolumn{4}{c}{\cellcolor[HTML]{CBCEFB}\textbf{Multimodal (N = 429431)}}                           \\ \hline
Early fusion avg pool & 56.35          & 46.72                & BU+Bert          \\ \hline
\multicolumn{1}{l}{Early fusion first token} & \multicolumn{1}{l}{56.98} & \multicolumn{1}{l}{48.34} & BU+Bert \\ \hline
Weighted late fusion & 65.14          & 60.27                & BU+Bert          \\ \hline
MMBT              & 66.33 & 62.24    & BU+Bert \\ \hline
VisualBERT                 & 66.03          & 61.47                & VinVL+Bert           \\ \hline
\end{tabular}
\caption{Performance comparison between image-only, text-only and multimodal models. Under features column (Feat), BU+Bert indicates bottom up features with 50 regions and token representations from BERT-base model. N indicates number of samples used for the modalities.
}
\vspace{-6mm}
\end{table}
\label{Table 1}

\begin{figure*}[h!]
    \begin{center}
    \includegraphics[scale=0.24]{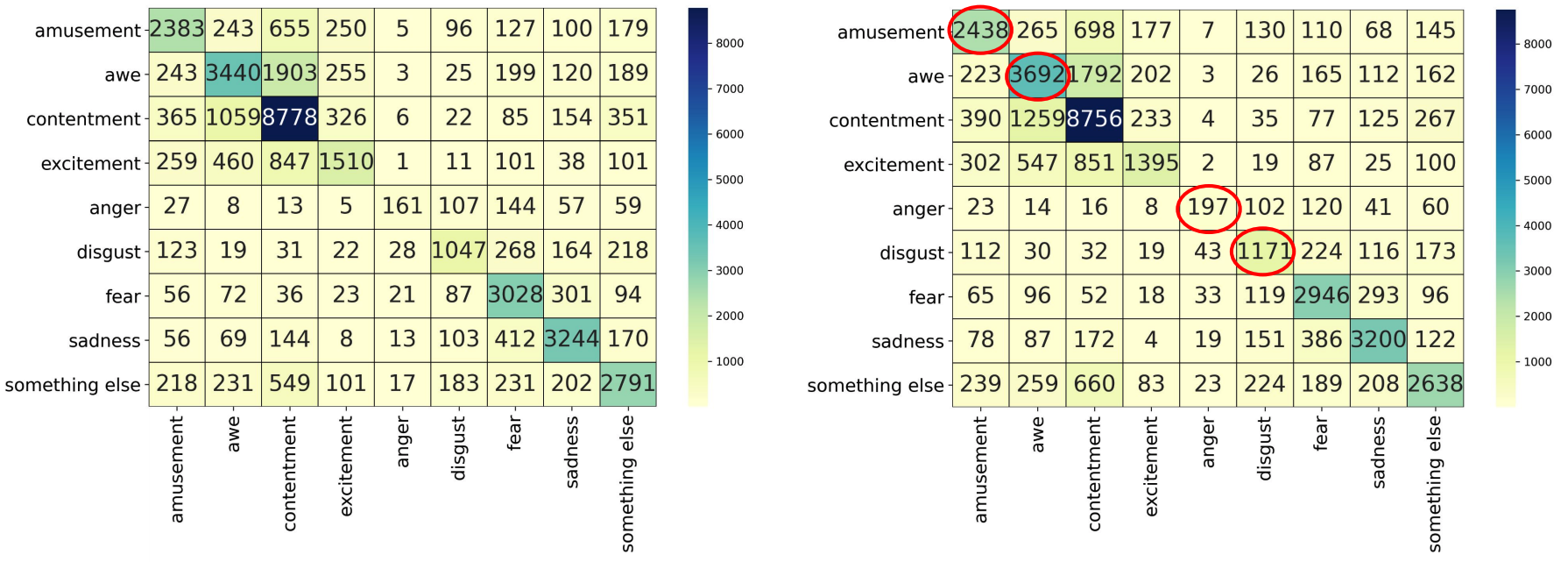}
    \caption{Confusion matrix plots for BERT (left) and MMBT (right). Red circles on the right confusion matrix plot signify emotion classes where MMBT performs better as compared to BERT.}
    \label{Confusion matrix comparison}
    \end{center}
\end{figure*}
\vspace{-2mm}
\subsubsection{Performance variation with image features}

We explore different configurations of bottom up and VinVL \cite{zhang2021vinvl} features by varying the number of detected boxes. From Table 2, it can be seen that there is a slight improvement on usage of bottom up features ($k = 50$), when compared with VinVL \cite{zhang2021vinvl}. However, the overall impact of varying the configuration of region-wise features does not result in significant improvements in emotion classification performance.

\begin{table}[h!]
\begin{center}
\begin{tabular}{|c|c|c|}
\hline
\textbf{Img} & \textbf{Acc.} & \textbf{F1} \\ \hline
BU-10        & 66.03        & 61.69       \\ \hline
BU-30        & 66.07        & 61.39       \\ \hline
BU-50 & {\color[HTML]{000000} 66.33} & {\color[HTML]{000000} 62.24} \\ \hline
VinVL-10     & 66.26        & 61.79       \\ \hline
VinVL-20     & 66.05        & 61.76       \\ \hline
\end{tabular}
\caption{Performance comparison(Accuracy and Macro-F1) for various input feature configurations for MMBT model. The text token representations are taken from pretrained BERT model. model-$k$: features from top-k bounding boxes sorted by scores from BU(bottom-up) and VinVL. }
\end{center}
\vspace{-6mm}
\end{table}


\subsection{Discussion}
\begin{figure*}[h!]
    \begin{center}
    \includegraphics[scale=0.24]{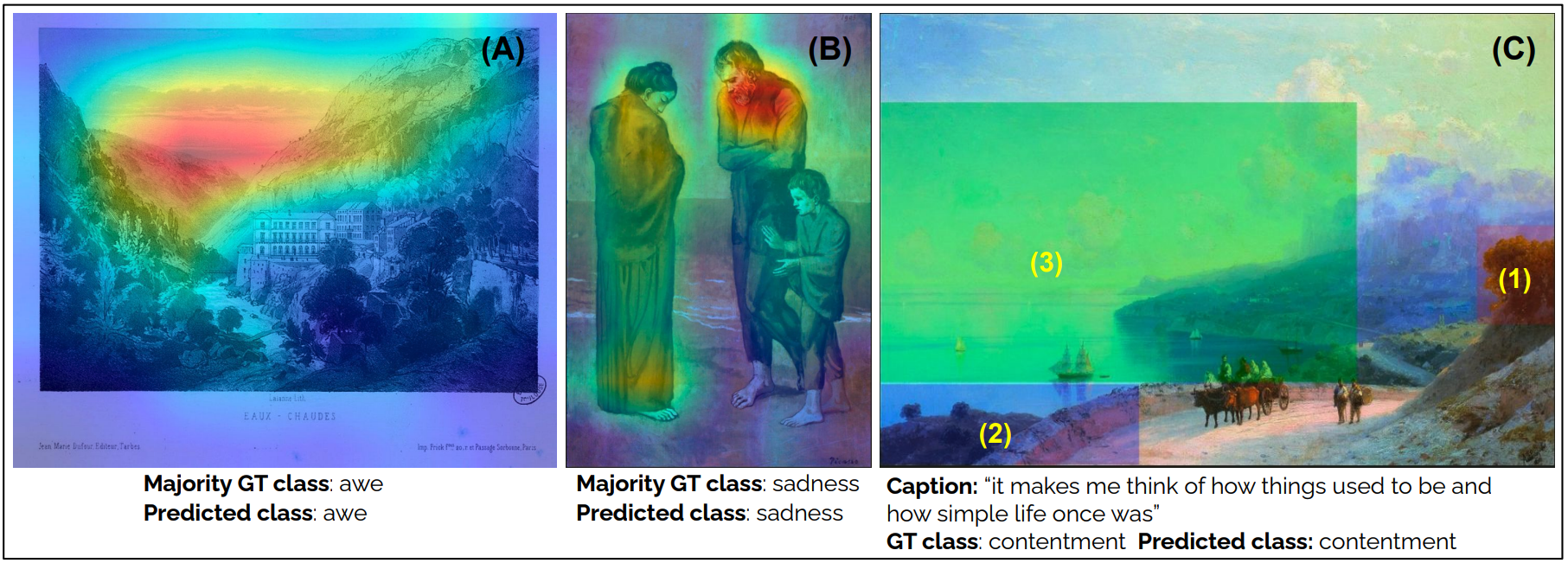}
    \caption{Visualization for image-only and multimodal models. Grad-cam visualizations shown for image-only model (VGG-16) making correct predictions for two cases involving (A) positive emotion (awe) and (B) negative emotion (sadness). Important image regions marked for MMBT model in order of increasing importance scores (1), (2), (3) for correct prediction.}
    \label{Visualization comparison}
    \end{center}
\end{figure*}
As seen from Table 1, the MMBT model (initialized with pretrained BERT weights) along with bottom up region features ($k$ = 50) performs on par with the text-only finetuned BERT baseline. Further, as seen in Fig.~\ref{Confusion matrix comparison}, MMBT shows an improvement in performance for extreme positive and negative emotion classes with overt visual features: amusement, awe, anger and disgust.  When compared with dual stream models that rely on explicit modeling of relationships between text tokens and region-wise features, single stream models like MMBT, VisualBERT also perform better. For image-only models, accuracies and F1-scores are reported based on the predicted dominant emotion label (in terms of occurrences). Their inferior performance can be attributed to the challenge involved in predicting a single emotion label from an art work due to inherent subjectivity and multiple possible interpretations.

\subsection{Visualizations}

We provide visualizations for image-only model i.e. VGG16 using Grad-CAM \cite{Selvaraju_2017_ICCV}. As shown in Fig \ref{Visualization comparison}(A), for ground truth majority class \enquote{awe}, the VGG-16 network is focusing on the mountains and the sky region while predicting the correct majority class. In case of \enquote{sadness} label shown in Fig \ref{Visualization comparison}(B), the network focuses more on the expression shown on the man's face, while predicting the correct label \enquote{sadness}. For the MMBT model, since we rely on region wise embeddings from images, we compute the importance of individual image regions using gradient based attributions. We normalize the importance of the image regions by dividing the individual importance scores by the maximum important score and showcase the top-3 important regions in the image, marked as (1), (2) and (3).

%



\section{Conclusions}
In this study, we explore multiple approaches for predicting evoked emotions from art-works through the perspective of different modalities. Based on the results, it can be seen that single-stream multimodal models like MMBT and VisualBERT perform better when compared to dual-stream approaches. In terms of visual information for multimodal analysis, including region-based features results in improvements over the text-only baseline i.e., BERT.  Future directions include extraction of holistic image features based on art-styles, color, lighting and modeling their interactions with the textual captions to classify evoked emotions. 



{\small
\bibliographystyle{ieee_fullname}
\bibliography{egbib}
}

\end{document}